\documentclass[10pt,twocolumn,letterpaper]{article}
\usepackage[pagenumbers]{paperstyle}
\usepackage{booktabs}
\usepackage{graphicx}
\usepackage{microtype}
\usepackage{amsmath,amssymb}
\usepackage{xcolor}
\usepackage{xspace}
\usepackage{multirow}
\usepackage{tabularx}
\usepackage{pifont}

\newcommand{\method}{OmniSenMo\xspace}
\definecolor{nabg}{HTML}{EBEBEB}
\newcommand{\no}{\cellcolor{nabg}--}

\definecolor{marktick}{HTML}{6A2FC9}
\newcommand{\cmark}{\textcolor{marktick}{\ding{51}}}
\newcommand{\xmark}{\textcolor{black!50}{\ding{55}}}
\newcommand{\pmark}{\textcolor{marktick}{$\circ$}}
\usepackage{cuted}
\usepackage{caption}
\newcommand{\raw}[1]{\,{\scriptsize(#1)}}
\usepackage{colortbl}
\usepackage{etoolbox}
\definecolor{oursbg}{HTML}{F1ECFB}
\newcommand{\oursrow}{\rowcolor{oursbg}}
\newcommand{\nobg}{\cellcolor{white}}
\newcolumntype{O}{>{\columncolor{oursbg}}c}
\newcommand{\tablestyle}{\setlength{\aboverulesep}{0pt}\setlength{\belowrulesep}{0pt}%
  \renewcommand{\arraystretch}{1.12}}
\AtBeginEnvironment{table}{\tablestyle}
\AtBeginEnvironment{table*}{\tablestyle}
\makeatletter
\renewcommand\paragraph{\@startsection{paragraph}{4}{\z@}{.5em \@plus 1pt}{-1em}{\normalfont\normalsize\bfseries}}
\makeatother
\definecolor{linkblue}{rgb}{0.21,0.49,0.74}
\usepackage[pagebackref,breaklinks,colorlinks,allcolors=linkblue]{hyperref}
\makeatletter

\let\AtPageUpperLeft\undefined

\let\AtTextUpperLeft\undefined

\let\ESO@Hline\undefined
\let\ESO@HookI\undefined
\let\ESO@HookIBG\undefined
\let\ESO@HookIFG\undefined
\let\ESO@HookII\undefined
\let\ESO@HookIIBG\undefined
\let\ESO@HookIIFG\undefined
\let\ESO@HookIII\undefined
\let\ESO@HookIIIBG\undefined
\let\ESO@HookIIIFG\undefined
\let\ESO@MEMOIR@corr\undefined
\let\ESO@Vline\undefined
\let\ESO@color\undefined
\let\ESO@colorbox\undefined
\let\ESO@div\undefined
\let\ESO@fcolorbox\undefined
\let\ESO@gap\undefined
\let\ESO@gridDelta\undefined
\let\ESO@gridDeltaY\undefined
\let\ESO@gridcolor\undefined
\let\ESO@griddelta\undefined
\let\ESO@griddeltaY\undefined
\let\ESO@gridlines\undefined
\let\ESO@gridunit\undefined
\let\ESO@gridunitname\undefined
\let\ESO@hline\undefined
\let\ESO@isMEMOIR\undefined
\let\ESO@labelfactor\undefined
\let\ESO@subgridcolor\undefined
\let\ESO@subgridlines\undefined
\let\ESO@subgridstyle\undefined
\let\ESO@tempdima\undefined
\let\ESO@tempdimb\undefined
\let\ESO@vline\undefined
\let\ESO@yoffsetI\undefined
\let\ESO@yoffsetII\undefined

\makeatother
\usepackage{pdfpages}
\makeatletter\let\@ShipoutPicture\relax\makeatother
\hypersetup{pdftitle={OmniSenMo: A Unified Multisensory Language Model for Generalizable Human Motion Estimation, Understanding, and Prediction}, pdfauthor={Cheng Guo, Qiming Cao, Pu (Perry) Wang, Pu Wang, Hongfei Xue}}
\makeatletter
\patchcmd{\@maketitle}{\vskip .375in}{\vskip .25in}{}{\errmessage{title space patch failed}}
\patchcmd{\@maketitle}{\vspace*{24pt}}{\vspace*{8pt}}{}{\errmessage{author space patch failed}}
\makeatother

\title{\method: A Unified Multisensory Language Model for Generalizable\\
Human Motion Estimation, Understanding, and Prediction}
\author{Cheng Guo$^{1}$ \quad Qiming Cao$^{2}$ \quad Pu (Perry) Wang$^{3}$ \quad Pu Wang$^{1}$ \quad
Hongfei Xue$^{1}$\thanks{Corresponding author.}\\
{\normalsize $^{1}$University of North Carolina at Charlotte \quad $^{2}$Purdue University \quad $^{3}$Mitsubishi Electric Research Laboratories (MERL)}\\[-2pt]
{\tt\small \{cguo3, pwang13, hxue2\}@charlotte.edu, cao393@purdue.edu, pwang@merl.com}\\[-3pt]
{\small Project page: \url{https://ch3ngguo.github.io/omnisenmo/}}}
\begin{document}
\maketitle
\begin{strip}
\vspace{-5em}
\centering
\noindent\makebox[\linewidth][c]{\includegraphics[width=0.88\linewidth]{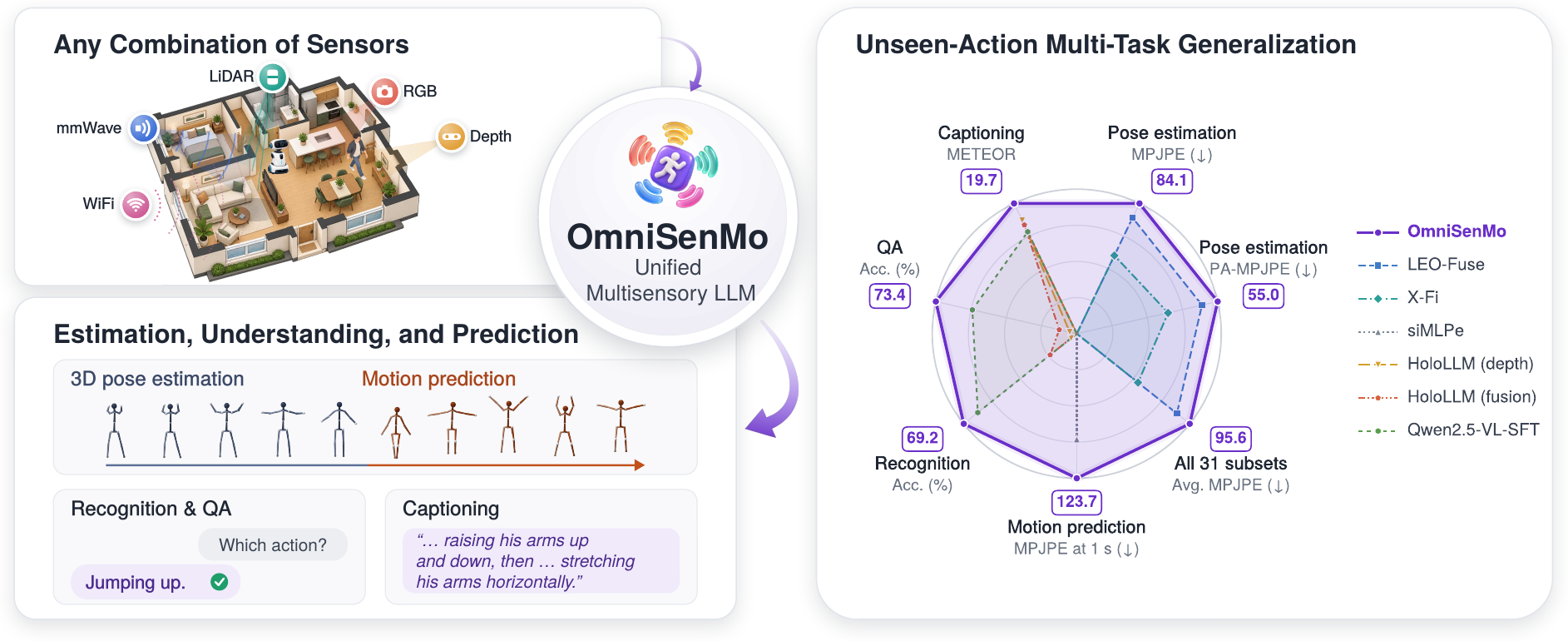}}
\captionsetup{hypcap=false}
\captionof{figure}{\textbf{A unified model for any sensor subset, generalizing to unseen actions.}
From any subset of the five sensors, the same \method model, pretrained on simulated sensors, estimates 3D human pose,
understands actions through recognition, question answering, and captioning, and predicts future motion, also for actions
absent from the real training data.
Right: unseen MM-Fi actions, where prior methods appear only on the tasks they support.}
\captionsetup{hypcap=true}
\label{fig:teaser}
\end{strip}

\begin{abstract}
Multimodal large language models (MLLMs) estimate, understand, and predict human motion, but mainly from RGB cameras.
Depth, LiDAR, mmWave, and WiFi complement RGB under poor lighting, through occlusions, and in privacy-sensitive applications, but existing multisensory models cover only some of these sensing tasks. Covering all of them from any sensor subset is challenging, since the tokens of heterogeneous sensors are semantically inconsistent and spatially misaligned, paired multisensory data are scarce, and the tasks need different temporal granularities. We present \method, a multisensory language model that takes the
human skeleton as a common reference for all sensors. Its skeleton-grounded motion representation turns each sensor into
joint tokens, one per body joint, so that any sensor subset is fused joint by joint and expressed in the motion
vocabulary that the language model learned from motion capture and text. We pretrain it on all five sensors generated by
our multisensory simulator, and Multi-Granularity Observation gives each task the frames it needs. On a cross-action split
of MM-Fi, the same model with all five sensors outperforms the models compared on each task, including LEO-Fuse in pose estimation (84.1 \vs
89.0\,mm MPJPE) and HoloLLM in unseen-action recognition (69.2\% \vs 42.8\%). Even without RGB, it outperforms both
methods that use RGB, and in a cross-dataset validation on XRF55, it recognizes unseen actions far better than HoloLLM.
\end{abstract}

\section{Introduction}

Embodied agents such as household robots increasingly share homes with people~\cite{kim2024openvla,li2024embodiedagent}.
To assist humans, an agent needs to know where they are, what they are doing, and how they will move next~\cite{puig2024habitat3}.
Vision-based multimodal large language models (MLLMs) combine visual perception with language reasoning~\cite{liu2023llava} to support human pose estimation, action understanding, and motion prediction~\cite{chen2025motionllm,feng2024chatpose,li2024unipose,wang2026unimotion,wang2026superman}, with recent models unifying multiple motion tasks and outperforming their single-task counterparts~\cite{wang2026unimotion,wang2026superman}.
Yet relying solely on RGB cameras limits these models in real-world settings.
For example, depth sensors complement RGB with direct depth measurements~\cite{shotton2011realtime}, while LiDAR, mmWave, and WiFi enable sensing under poor lighting~\cite{yang2024mmfi,zhou2025holollm}.
Moreover, mmWave and WiFi enable sensing through occlusions and even walls~\cite{li2019rfaction} and support privacy-sensitive applications by not recording visual appearance~\cite{yang2024mmfi,aslan2026leofuse}.
These complementary capabilities have motivated multisensory human sensing~\cite{yang2024mmfi,chen2024xfi}.

\begin{table}[t]
\centering
\caption{\textbf{Capabilities of related models.}  I: RGB, D: depth, L: LiDAR, R: mmWave, W: WiFi.  Any
subset: any nonempty sensor subset.  Unified: a single model for all tasks.
Est.: estimation, HAR: action recognition, QA: question answering, Cap.: action captioning, Pred.: prediction.  \pmark: one sensor per input, fusion only in its appendix.}
\label{tab:capability}
\footnotesize
\setlength{\tabcolsep}{2pt}
\resizebox{\columnwidth}{!}{%
\begin{tabular}{l*{12}{c}}
\toprule
 & \multicolumn{5}{c}{Sensors} & & \multicolumn{5}{c}{Tasks} & \\
\cmidrule(lr){2-6}\cmidrule(lr){8-12}
Method & I & D & L & R & W & \shortstack{Any\\subset} & Est. & HAR & QA & Cap. & Pred. & Unified \\
\midrule
X-Fi~\cite{chen2024xfi} & \cmark & \cmark & \cmark & \cmark & \cmark & \cmark & \cmark & \cmark & \xmark & \xmark & \xmark & \xmark \\
LEO-Fuse~\cite{aslan2026leofuse} & \cmark & \cmark & \cmark & \cmark & \cmark & \cmark & \cmark & \cmark & \xmark & \xmark & \xmark & \xmark \\
HoloLLM~\cite{zhou2025holollm} & \cmark & \cmark & \cmark & \cmark & \cmark & \pmark & \xmark & \cmark & \cmark & \cmark & \xmark & \cmark \\
mmPred~\cite{fan2026mmpred} & \xmark & \xmark & \xmark & \cmark & \xmark & \xmark & \xmark & \xmark & \xmark & \xmark & \cmark & \cmark \\
UniPose~\cite{li2024unipose} & \cmark & \xmark & \xmark & \xmark & \xmark & \xmark & \cmark & \xmark & \xmark & \xmark & \xmark & \cmark \\
Superman~\cite{wang2026superman} & \cmark & \xmark & \xmark & \xmark & \xmark & \xmark & \cmark & \xmark & \xmark & \xmark & \cmark & \cmark \\
UniMotion~\cite{wang2026unimotion} & \cmark & \xmark & \xmark & \xmark & \xmark & \xmark & \cmark & \xmark & \xmark & \cmark & \cmark & \cmark \\
\midrule
\oursrow\method & \cmark & \cmark & \cmark & \cmark & \cmark & \cmark & \cmark & \cmark & \cmark & \cmark & \cmark & \cmark \\
\bottomrule
\end{tabular}}
\vspace{-1em}
\end{table}

Assisting humans requires multisensory models that support pose estimation and motion prediction, alongside action understanding through recognition, action question answering (QA), and captioning.
These models should also generalize to \emph{unseen actions} encountered in everyday life but absent from their real training data.
Existing multisensory fusion methods, such as X-Fi~\cite{chen2024xfi} and LEO-Fuse~\cite{aslan2026leofuse}, support pose estimation and action recognition from arbitrary sensor subsets (Tab.~\ref{tab:capability}).
However, each task requires a separately trained model, and action recognition remains limited to predefined labels.
Sensor--language models enable richer action understanding, including QA and captioning, by connecting mmWave~\cite{lai2026radarllm,yan2025mmexpert,zhang2026mmmind}, WiFi~\cite{zhang2025wichat}, and multiple sensing modalities~\cite{zhou2025holollm} to language models, but they do not estimate pose or predict motion.
Moreover, existing multisensory fusion models are evaluated only on seen actions, leaving their generalization to unseen actions unexplored.
We therefore ask the question illustrated in Fig.~\ref{fig:teaser}: \emph{Can a single language model estimate, understand, and predict human motion from any available sensor subset, including actions absent from its real training data?}
To answer this question, three challenges concerning sensor heterogeneity, data scarcity, and diverse task requirements must be addressed.

First, sensor heterogeneity creates semantic inconsistency and spatial misalignment between sensor tokens.
Semantically, sensor-specific encoders emphasize different aspects of motion and entangle them with sensor-dependent patterns unrelated to motion, such as image backgrounds or environment-dependent multipath.
Spatially, sensor tokens can differ in number and lack correspondence to the same body joints~\cite{zhou2025holollm}.
Inspired by shared motion--language representations~\cite{jiang2023motiongpt,luo2024m3gpt}, we identify the human skeleton as a common reference for aligning heterogeneous observations of the same human motion.
Instead of the learnable queries~\cite{li2023blip2,han2024onellm} or pooled CLIP tokens~\cite{radford2021clip,zhou2025holollm} that existing multimodal language models receive, we therefore propose a unified skeleton-grounded motion representation in which each sensor produces one joint token per joint and frame, supervised by the 3D position of that joint and its discrete motion token in a shared, pretrained motion vocabulary.
These shared targets encourage consistent, joint-focused motion semantics across sensors, while explicit joint correspondence provides spatial alignment.
The resulting representation is sensor-agnostic in its joint semantics, enabling joint-wise fusion of any available sensor subset and reuse of motion knowledge learned from motion capture and text.
We further pretrain the model to recover masked sensors' joint tokens from the remaining sensors rather than reconstructing image patches~\cite{bachmann2022multimae} or generic modality-specific features~\cite{ma2021smil,woo2023actionmae}, encouraging it to exploit complementary joint information when sensors are unavailable.

Second, paired multisensory data are scarce.
Collecting calibrated, synchronized sensor observations is costly and time-consuming, so existing multisensory datasets involving radio-frequency (RF) signals~\cite{chen2022mmbody,yang2024mmfi,wang2024xrf55,jiang2025cuhkx} cover far fewer actions and subjects than RGB or motion-capture datasets~\cite{kay2017kinetics,mahmood2019amass}.
Yet both sensor--language alignment~\cite{han2024onellm,zhou2025holollm,dai2025babel} and joint multisensory pretraining~\cite{bachmann2022multimae,mizrahi20234m,bachmann2024fourm21,narayanswamy2025scaling} require large paired corpora.
To mitigate this scarcity, we build a multisensory simulator that generates synchronized observations from all five sensors for diverse motion-capture sequences~\cite{mahmood2019amass}, paired with 3D pose and
language~\cite{punnakkal2021babel,guo2022humanml3d}, enabling joint pretraining before adaptation to real data.
Besides simulation, we
also distill the vision encoder of a vision--language model (VLM) into non-RGB encoders through feature matching on synchronized frames.

Third, the tasks require different temporal granularities~\cite{wen2023htt}: local motion detail for estimation and prediction, and broader action context for understanding.
Dense full-action inputs could provide both, but token counts grow with sensors and frames.
Under a limited frame budget~\cite{li2024llamavid,markhorst2026figmo}, contiguous observations preserve local detail but may miss the action context needed for understanding.
We therefore introduce Multi-Granularity Observation (MGO), enabling the model to process inputs at different temporal granularities: contiguous frames for estimation, samples spanning an action for understanding, and recent history for prediction.

Together, simulated motion diversity, VLM knowledge transfer, and skeleton-grounded representations enable \method to reuse motion--language knowledge across sensors and generalize to unseen actions. Our contributions are:

\begin{itemize}
    \item We propose \method, to our knowledge the first unified model that estimates, understands, and predicts
    human motion from any sensor subset.
    Its skeleton-grounded motion representation aligns heterogeneous sensors through the human skeleton, since the joint
    tokens of every sensor correspond to the same joints and are supervised by the 3D position and motion token of each
    joint. Multi-Granularity Observation processes each task at its temporal granularity.
    \item We build a multisensory simulator that generates a corpus of five sensors from annotated motion capture, paired
    with 3D pose and language, and pretrain \method on this corpus before adapting it to real data, which mitigates data
    scarcity and improves generalization to unseen actions.
    \item On a cross-action split of MM-Fi, \method with all five sensors outperforms the models compared on each task, and in a cross-dataset
    validation on XRF55 it recognizes unseen actions far better than HoloLLM.  We will release the code and the simulated
    corpus.
\end{itemize}

\section{Related Work}
\label{sec:related}

\begin{figure*}[t]
\centering
\includegraphics[width=\textwidth]{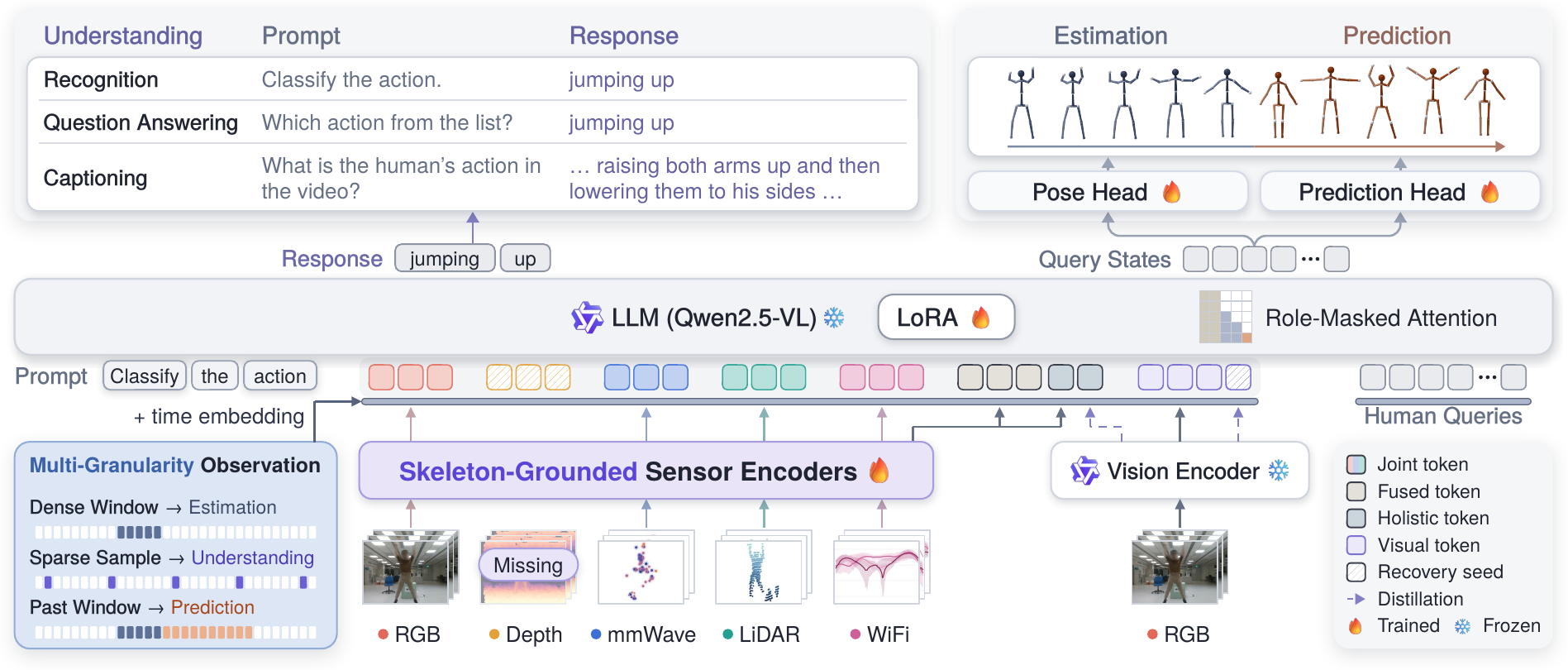}
\caption{\textbf{Overview of \method.}  Each sensor of any subset passes through its skeleton-grounded
encoder, which outputs one joint token per joint and frame, supervised by the 3D position and the motion token of that
joint.  Learned recovery seeds take the place of a missing sensor (here depth).
One context of the language model (Qwen2.5-VL with LoRA)
contains the prompt, the joint, fused, holistic, and visual tokens, and the human queries under
role-masked attention.  In training, the frozen vision encoder also serves as the VLM teacher (dashed).
Multi-Granularity Observation (MGO) sets the frames that each task uses.  Query states feed the
pose and prediction heads, and the response gives recognition, QA, and captions.  The estimated window is shown in
blue and its future motion in orange.}
\label{fig:pipeline}
\end{figure*}

\noindent\textbf{Multisensory human sensing.}
Beyond RGB cameras, human pose has been estimated from depth~\cite{shotton2011realtime},
LiDAR~\cite{ye2024lpformer,an2025dapt}, radar~\cite{zhao2018rfpose,zhao2018rfpose3d,fan2024mmdiff}, and
WiFi~\cite{wang2019personinwifi,zhou2023metafipp,yan2024personinwifi3d}, and human motion has been predicted from
mmWave~\cite{fan2026mmpred}.  With the emergence of paired multisensory
datasets~\cite{chen2022mmbody,yang2024mmfi,wang2024xrf55,jiang2025cuhkx}, X-Fi handles arbitrary sensor subsets for pose
estimation and action recognition without training a model per combination~\cite{chen2024xfi}, and
LEO-Fuse applies one task-general architecture to several tasks but trains a separate model for
each~\cite{aslan2026leofuse}, while AWC balances the learning of RGB, LiDAR, mmWave, and WiFi for pose
estimation~\cite{qi2026awc}.  For actions absent from training, zero-shot
recognition methods align mmWave, WiFi, or IoT signals with text~\cite{cao2024mmclip,shen2026zerofi,zhou2026tent}, and
DeSPITE embeds LiDAR point clouds with skeletons, IMU, and text~\cite{kreutz2025despite}.
 HoloLLM integrates five sensors into an MLLM for action recognition, QA,
and captioning~\cite{zhou2025holollm}, while language models for
mmWave~\cite{lai2026radarllm,yan2025mmexpert,shin2026mmwaveqa,zhang2026mmmind} and WiFi~\cite{zhang2025wichat} handle a single sensor.  
\method estimates,
understands, and predicts human motion in a unified model for any sensor subset.

\noindent\textbf{Multimodal and motion--language models.}
MLLMs add new modalities through modality-specific encoders~\cite{xu2024pointllm,su2023pandagpt} or one
frozen encoder shared across modalities~\cite{han2024onellm,zhang2023metatransformer}.
For human motion, discrete motion
tokens support generation~\cite{zhang2023t2mgpt,guo2024momask,shah2026walk}, and
language models generate and interpret
motion~\cite{jiang2023motiongpt,luo2024m3gpt,chen2025motionllm}.  Timestamped
skeleton tokens let them reason about order and duration~\cite{markhorst2026figmo}.
ChatPose~\cite{feng2024chatpose}, UniPose~\cite{li2024unipose}, and UniMotion~\cite{wang2026unimotion} unify pose or
motion understanding and generation with text and images, and
Superman~\cite{wang2026superman} adds video-based 3D pose estimation and motion
prediction.
These motion models observe humans only through RGB or motion capture.

\noindent\textbf{Data scarcity and simulation.}
Paired sensor data are scarce, so sensing models often borrow supervision from RGB.  Cross-modal distillation trains
depth, audio, or radio models with vision teachers on synchronized RGB~\cite{gupta2016cross,aytar2016soundnet,zhao2018rfpose},
and binding methods align further modalities with images, language, or modalities already
aligned~\cite{girdhar2023imagebind,yang2024binding,zhu2024languagebind,wang2025omnibind,dai2025babel}.  Simulation instead creates paired data.
Synthetic humans are used to train visual human pose and shape
estimators~\cite{varol2017surreal,patel2021agora,black2023bedlam,tesch2025bedlam2}.  Sensor simulators
have converted motion capture, videos, or generated scenes into depth images~\cite{shotton2011realtime}, inertial
signals~\cite{huang2018dip,kwon2020imutube}, WiFi signals~\cite{cai2020teaching}, and Doppler and mmWave
signals~\cite{ahuja2021vid2doppler,deng2023midas,zhang2022synmotion,xue2023mmgpe,chen2023rfgenesis} to train sensing
models.  A few simulate two or three sensors together, such as LiDAR with inertial sensors~\cite{ren2023lip} or LiDAR
with RGB cameras~\cite{pan2025licampose}.  More recent work pretrains LiDAR and mmWave models on simulated data before
adapting them to real data~\cite{an2025dapt,cao2024mmclip,guo2026mmsimprior}, or synthesizes sensor--text data to align
inertial signals with text~\cite{zhang2024unimts} and to train language models for
mmWave~\cite{lai2026radarllm,yan2025mmexpert}.
None of these simulators pairs mmWave or WiFi with cameras or LiDAR.  To our knowledge, we are the first to simulate all
five sensors from the same human motions and to pretrain a multisensory model on them jointly.

\section{Method}
\label{sec:method}

Let $\mathcal{M}=\{\mathrm{I},\mathrm{D},\mathrm{L},\mathrm{R},\mathrm{W}\}$ denote the five sensors (RGB, depth,
LiDAR, mmWave, and WiFi), and let $S\subseteq\mathcal{M}$ be the nonempty subset available for a sample.  Each sensor
$m\in S$ observes the subject over $T$ frames $\mathbf{x}^m_1,\dots,\mathbf{x}^m_T$, which we call its observation.  We
represent the human at frame $t$ by $J{=}17$ joints $\mathbf{X}_t\in\mathbb{R}^{J\times3}$, decomposed into the pelvis
trajectory $\mathbf{p}_t$ and the local pose $\mathbf{P}_t=\mathbf{X}_t-\mathbf{1}\mathbf{p}_t^{\top}$.  A frozen vector-quantized (VQ)
tokenizer~\cite{vandenoord2017vqvae} learned on AMASS~\cite{mahmood2019amass} quantizes each joint $j$ of $\mathbf{P}_t$
into a discrete motion token $u_{t,j}$, while the trajectory stays continuous.
From any
$S$, \method estimates the 3D human pose $\mathbf{X}_t$ of every frame, understands the action through action recognition, action
question answering, and captioning, and predicts future motion, all with a unified multisensory language model.
As illustrated in Fig.~\ref{fig:pipeline}, \method comprises skeleton-grounded sensor encoders, which map every sensor
into a unified representation so that the model can fuse any sensor subset and recover missing sensors
(Sec.~\ref{sec:interface}), and a language model, which serves tasks of different temporal granularity through
MGO (Sec.~\ref{sec:mohq}).  To mitigate data scarcity, multisensory
sim-to-real pretraining trains the whole model on fully paired simulated data before adapting it to real data
(Sec.~\ref{sec:simdata}).

\subsection{Skeleton-grounded joint tokens}
\label{sec:interface}

To address the semantic inconsistency and spatial misalignment of sensor tokens, we ground every sensor in the human
skeleton, whereas sensor--language models ground each sensor in
language~\cite{zhou2025holollm,lai2026radarllm,zhang2026mmmind}.  Each sensor is encoded into $J$ joint tokens per frame,
each supervised by the 3D position and the motion token of its joint.  Unlike raw sensor signals, human motion
can be learned by a language model as a foreign language, as in motion--language
models~\cite{jiang2023motiongpt,luo2024m3gpt,li2024unipose}.

\noindent\textbf{Skeleton-grounded encoders.}
Each encoder builds on a backbone pretrained on simulated data, so that it already has the capacity to represent human
motion before it encounters real-world data.  One learned embedding per joint cross-attends to the backbone features of every
frame, and a temporal Transformer turns the results into joint features $\mathbf{z}^m_{t,j}$.
An MLP regresses the
local position $\hat{\mathbf{P}}^m_{t,j}$ of each joint, and a linear classifier predicts its motion token $u_{t,j}$ as a
distribution $\hat{\mathbf{u}}^m_{t,j}$ over the VQ codebook.
With the pelvis position $\hat{\mathbf{p}}^m_t$ regressed from the global
backbone features,
each encoder is trained with the grounding loss
\begin{equation}
  \begin{aligned}
  \mathcal{L}_{\mathrm{skel}}^{m}={}&\operatorname{CE}(\hat{\mathbf{u}}^m,u)
  +\lambda_{\mathrm{loc}}\,e(\hat{\mathbf{P}}^m,\mathbf{P})\\
  &+\lambda_{\mathrm{root}}\,e(\hat{\mathbf{p}}^m,\mathbf{p})
  +\lambda_{\mathrm{abs}}\,e(\hat{\mathbf{X}}^m,\mathbf{X}),
  \end{aligned}
  \label{eq:skel}
\end{equation}
where $\operatorname{CE}$ is the cross-entropy, $e$ is the mean Euclidean distance over frames and joints,
$\hat{\mathbf{X}}^m_t=\hat{\mathbf{P}}^m_t+\mathbf{1}(\hat{\mathbf{p}}^m_t)^{\top}$, and the $\lambda$ are loss weights.  These shared
targets give the joint tokens of all sensors consistent motion semantics.  A projector maps every frame to $J$
joint tokens, and non-RGB encoders also pool two holistic tokens $\mathbf{w}^m$ per observation.

\begin{figure*}[t]
\centering
\includegraphics[width=\textwidth]{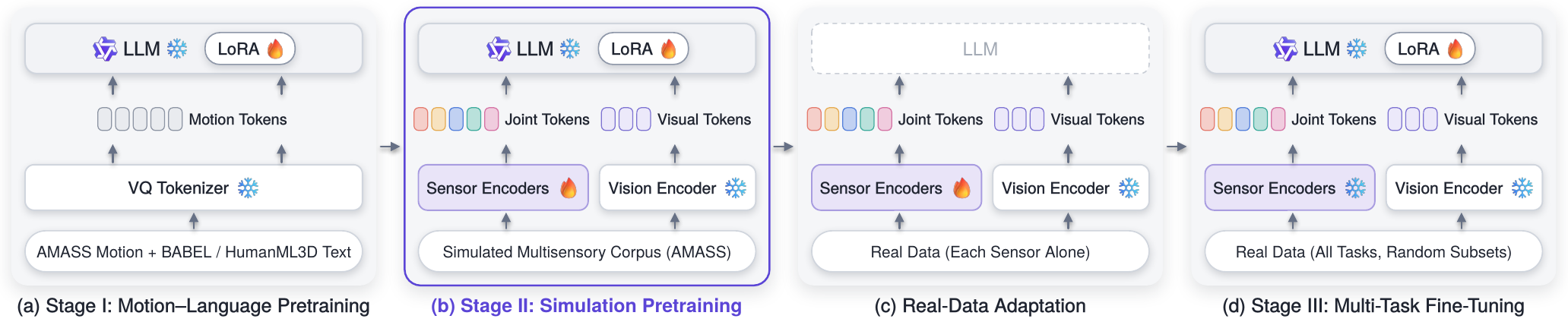}
\caption{\textbf{From simulation to real data} (Sec.~\ref{sec:simdata}): trained (flame) and frozen
(snowflake) parts in each stage.  Dashed parts are not used.
The fusion module, human queries, and heads (Fig.~\ref{fig:pipeline}) are trained together with LoRA.  Stage~II trains on the simulated multisensory corpus, built from the AMASS motions of Stage~I.  On XRF55, Stage~III also
tunes part of each encoder (supplementary material).}
\label{fig:training}
\end{figure*}

\noindent\textbf{Joint correspondence.}
Every encoder outputs its joint tokens in the same fixed joint order, and token $j$ is supervised by joint $j$, so the
$j$-th tokens of all sensors describe the same joint,
such as the left wrist, whether they come from RGB or from WiFi.  This removes their spatial misalignment, so the model can
combine the sensors joint by joint.  The fused tokens attend to the tokens of every available sensor and carry the per-sensor poses averaged with
per-joint weights normalized over $S$.  As the reliability of a sensor varies across joints~\cite{aslan2026leofuse},
these weights are supervised by the pose error of each sensor.

\noindent\textbf{Masked sensor recovery.}
Since all sensors share one token space aligned with the skeleton, a missing sensor leaves only a missing block in this
space.  Each sensor has learned recovery seeds
that take the place of its joint tokens in the context of the language model.  We mask
sensors at random, and the seeds of a masked sensor $m$
regress the features that the encoder computes from the masked input,
\begin{equation}
  \begin{aligned}
  \mathcal{L}_{\mathrm{rec}}^{m}={}&\ell(\tilde{\mathbf{z}}^m,\mathbf{z}^m)
  +\lambda_g\,\ell(\tilde{\mathbf{g}}^m,\mathbf{g}^m)\\
  &+\lambda_p\operatorname{MSE}(\tilde{\mathbf{p}}^m,\hat{\mathbf{p}}^m),
  \end{aligned}
  \label{eq:jm}
\end{equation}
where $\mathbf{z}^m$ are the joint features, $\mathbf{g}^m$ is their mean over joints, $\hat{\mathbf{p}}^m$ is the encoder's pelvis
position, tildes denote predictions from the seeds, and $\ell$ is the MSE between layer-normalized features.  
At inference, the same seeds replace
every missing sensor, as shown for depth in Fig.~\ref{fig:pipeline}.

\subsection{Unified language model for multi-granularity tasks}
\label{sec:mohq}

The tasks differ in their outputs and in their temporal granularity, yet \method serves all of them from any sensor
subset with the same language model and context.  We employ Qwen2.5-VL-7B~\cite{bai2025qwen25vl} as the language model,
frozen and adapted with LoRA~\cite{hu2022lora}.  Its context contains the prompt, the joint tokens of every sensor, with
recovery seeds in place of missing ones, the fused and holistic tokens of Sec.~\ref{sec:interface}, the visual tokens of the
frozen vision encoder, and the \emph{human queries}, followed by the response.  Given the task prompt, the
response answers recognition and QA and writes captions, while the human queries give the poses for estimation and
prediction.
The human queries contain one query for the pelvis and each joint of every frame and two queries for the
whole observation.  They are indexed by the same joints and frames as the joint tokens, so each query can attend to its
joint in every sensor.  The pose head regresses the pose, and the prediction head predicts
future motion.  Attention is role-masked as in UniPose~\cite{li2024unipose}, with
bidirectional attention among the human queries and causal attention for the response, and both attend to the
joint tokens.

\noindent\textbf{Multi-Granularity Observation (MGO).}
To handle the different temporal granularities of the tasks, MGO lets every task use the same joint tokens on its own
frames, chosen by one of three observation protocols,
\begin{equation}
  \begin{aligned}
  \mathcal{F}_{\mathrm{d}}&=\{t,\dots,t{+}T{-}1\}, &
  \mathcal{F}_{\mathrm{p}}&=\{t{-}T{+}1,\dots,t\},\\
  \mathcal{F}_{\mathrm{s}}&=\{f_i\in\mathcal{I}_i\}_{i=1}^{T}.
  \end{aligned}
  \label{eq:protocols}
\end{equation}
As shown in Fig.~\ref{fig:pipeline}, estimation takes the \emph{dense window} $\mathcal{F}_{\mathrm{d}}$ of contiguous frames.  Understanding takes the
\emph{sparse sample} $\mathcal{F}_{\mathrm{s}}$, with one frame $f_i$ from each of the $T$ equal segments $\mathcal{I}_i$
of the annotated action interval, as in HoloLLM~\cite{zhou2025holollm}.  Motion prediction takes the \emph{past window}
$\mathcal{F}_{\mathrm{p}}$ and predicts the frames that follow without future sensor data.
Every joint token and every per-frame human query carries a time embedding of its frame, so the same human queries
estimate the pose at observed frames and predict it at future frames.

\subsection{Multisensory sim-to-real pretraining}
\label{sec:simdata}
\label{sec:sim2real}

\begin{table*}[t]
\centering
\caption{\textbf{Main results on unseen MM-Fi actions}, the same \method model for every task.  Sensors are
given in parentheses (all five otherwise).
(a)~Avg$_{31}$: MPJPE averaged over all 31 sensor subsets.  (b)~Calibrated top-1 (\%), uncalibrated in parentheses.
(c)~MPJPE averaged over the next 1\,s (Avg) and at 1\,s, from \method's estimated past (upper) or from mmWave
(lower, HumanMAC from mmPred's estimate).  Bold: best in each block.  --: not available.  $^\dagger$Our
reimplementation.  $^\S$Weighted-sum fusion.}
\label{tab:main}
\footnotesize
\setlength{\tabcolsep}{2pt}
\begin{minipage}[t]{0.396\textwidth}
\centering
\textbf{(a) 3D pose estimation (mm)}\\[2pt]
\begin{tabular}{lcccc}
\toprule
Method & MPJPE$\downarrow$ & PA$\downarrow$ & MRPE$\downarrow$ & Avg$_{31}{\downarrow}$ \\
\midrule
X-Fi~\cite{chen2024xfi} & 105.1 & 66.7 & 75.9 & 123.9 \\
LEO-Fuse$^\dagger$~\cite{aslan2026leofuse} & 89.0 & 58.2 & 67.3 & 101.3 \\
\oursrow\method & \textbf{84.1} & \textbf{55.0} & \textbf{64.6} & \textbf{95.6} \\
\midrule
AWC$^\dagger$~\cite{qi2026awc} (I+L+R+W) & 91.8 & 61.0 & 66.4 & \no \\
\oursrow\method\ (I+L+R+W) & \textbf{86.5} & \textbf{55.2} & \textbf{65.6} & \no \\
\midrule
DAPT~\cite{an2025dapt} (L) & 90.6 & 60.8 & 74.8 & \no \\
\oursrow\method\ (L) & \textbf{88.9} & \textbf{55.8} & \textbf{70.3} & \no \\
\bottomrule
\end{tabular}
\end{minipage}\hfill
\begin{minipage}[t]{0.377\textwidth}
\centering
\textbf{(b) Motion understanding}\\[2pt]
\begin{tabular}{lccc}
\toprule
Method & HAR$\uparrow$ & QA$\uparrow$ & METEOR$\uparrow$ \\
\midrule
HoloLLM$^\S$~\cite{zhou2025holollm} & 40.8\raw{32.3} & 39.7\raw{32.5} & 18.1 \\
HoloLLM (I+R+W)$^\S$ & 42.8\raw{32.0} & 41.3\raw{32.2} & 18.1 \\
\oursrow\method & \textbf{69.2}\raw{61.7} & \textbf{73.4}\raw{56.9} & \textbf{19.7} \\
\midrule
Qwen2.5-VL-ZS (I) & 42.7\raw{36.1} & 42.3\raw{25.8} & 7.6 \\
Qwen2.5-VL-SFT (I) & 64.9\raw{61.4} & \textbf{63.9}\raw{62.9} & 17.6 \\
HoloLLM (I) & 31.5\raw{32.3} & 31.1\raw{32.5} & 18.4 \\
\oursrow\method\ (I) & \textbf{68.2}\raw{62.4} & 55.6\raw{53.2} & \textbf{19.9} \\
\bottomrule
\end{tabular}
\end{minipage}\hfill
\begin{minipage}[t]{0.212\textwidth}
\centering
\textbf{(c) Motion prediction (mm)}\\[2pt]
\begin{tabular}{lcc}
\toprule
Method & Avg$\downarrow$ & 1\,s$\downarrow$ \\
\midrule
siMLPe~\cite{guo2023simlpe} & 115.4 & 143.0 \\
AuxFormer~\cite{xu2023auxformer} & 121.2 & 150.5 \\
HumanMAC~\cite{chen2023humanmac} & 121.6 & 151.4 \\
\oursrow\method & \textbf{112.8} & \textbf{123.7} \\
\midrule
mmPred~\cite{fan2026mmpred} (R) & 132.3 & 154.1 \\
HumanMAC (R) & 136.3 & 159.1 \\
\oursrow\method\ (R) & \textbf{125.9} & \textbf{133.8} \\
\bottomrule
\end{tabular}
\end{minipage}
\end{table*}

\noindent\textbf{Simulated multisensory corpus.}
To mitigate the scarcity of paired multisensory data, we simulate every sensor from the SMPL-X~\cite{pavlakos2019smplx} mesh sequences of the
AMASS~\cite{mahmood2019amass} motions, which BABEL~\cite{punnakkal2021babel} and HumanML3D~\cite{guo2022humanml3d}
annotate with actions and text.
RGB and depth are rasterized from the mesh, LiDAR casts rays against it, and the mmWave and WiFi signals sum the paths
scattered by every mesh triangle.
Placed as in MM-Fi, the virtual sensors observe every sequence at the same frames.  Each simulated motion thus becomes
five synchronized sensor streams, paired by construction with the 3D joints regressed from the mesh and with its action
labels and captions.

\noindent\textbf{Distillation from the VLM teacher.}
Simulation supervises the encoders with 3D pose.  To transfer the generalization that a VLM gains from large and
diverse images, we distill the non-RGB encoders from a VLM teacher, the frozen vision encoder of the language model, computed
on the synchronized RGB of the same frames.  Let $\mathbf{y}_t$ be the normalized teacher feature of frame $t$ and
$\bar{\mathbf{y}}$ its mean over the observation, and let $\mathbf{s}^m_t$ be the frame features of encoder $m$ and
$\bar{\mathbf{w}}^m$ the mean of its holistic tokens.
With a linear projection $\mathbf{A}^m$,
\begin{equation}
  \mathcal{L}_{\mathrm{KD}}^{m}=\frac{1}{T}\sum_{t=1}^{T}d\bigl(\mathbf{A}^m\mathbf{s}^m_t,\mathbf{y}_t\bigr)
  +d\bigl(\mathbf{A}^m\bar{\mathbf{w}}^m,\bar{\mathbf{y}}\bigr),
  \label{eq:distill}
\end{equation}
where $d(\mathbf{a},\mathbf{b})=1-\cos(\mathbf{a},\mathbf{b})$.  The first term matches every frame and the second the
whole observation.  The RGB recovery seeds predict the same teacher features as $\tilde{\mathbf{y}}_t$, with
$\mathcal{L}_{\mathrm{rec}}^{\mathrm{VLM}}=\frac1T\sum_t d(\tilde{\mathbf{y}}_t,\mathbf{y}_t)$.  Neither loss uses action labels.

\noindent\textbf{Pretraining.}
Training moves from motion to simulated sensors to real data, as shown in Fig.~\ref{fig:training}.  In
\emph{Stage~I}, before any sensor is attached, the motion tokens extend the vocabulary of the language model, which
learns them with masked motion modeling~\cite{guo2024momask,chen2023humanmac} on AMASS and translation between motion
and text on BABEL~\cite{punnakkal2021babel} and HumanML3D~\cite{guo2022humanml3d}, so that it processes pose and text
together.  \emph{Stage~II} then pretrains the whole model on the simulated corpus.  Each encoder is first trained alone
with the grounding loss $\mathcal{L}_{\mathrm{skel}}$ and, for non-RGB sensors,
$\mathcal{L}_{\mathrm{KD}}$.  The whole model then learns from randomly masked sensor subsets.
As every simulated sensor observes the same labeled pose, the pose estimated from any subset is supervised by it, and
the recovery loss $\mathcal{L}_{\mathrm{rec}}$ takes the masked sensors themselves as targets.  The
RGB recovery seeds learn the teacher features with $\mathcal{L}_{\mathrm{rec}}^{\mathrm{VLM}}$, and the first term of
$\mathcal{L}_{\mathrm{KD}}$ keeps distilling the non-RGB encoders.  The model also learns to predict future motion and,
through the language model, to describe the motion with the action labels of BABEL and the captions of HumanML3D.

\noindent\textbf{Sim-to-real adaptation.}
We treat the sim-to-real gap as a gap in the sensor signals rather than in the motions or their language.
\emph{Real-data adaptation} therefore trains each encoder alone on the real training split with the same encoder losses,
so that it maps real signals onto the skeleton, while the projector, the fusion module, and the language model keep what they
learned in simulation.  Because every encoder produces the same joint tokens from simulated and real data, the motions
learned in simulation carry over to real data.  \emph{Stage~III} then fine-tunes the model on the real training actions
with the encoders frozen, one task per update and a random sensor subset per sample.  Network and training details are in the
supplementary material.

\section{Experiments}
\label{sec:experiments}

\subsection{Setup}
\label{sec:setup}

\noindent\textbf{Datasets and splits.}
Our main benchmark is a cross-action split of MM-Fi~\cite{yang2024mmfi}, which pairs five sensors
(I: RGB, D: depth, L: LiDAR, R: mmWave, W: WiFi) with 3D human pose.  We train on 21 of its actions and test on the other
six, as in its official code.
XRF55~\cite{wang2024xrf55} is an indoor action recognition dataset without 3D pose.  On the subset with public Kinect RGB
that HoloLLM~\cite{zhou2025holollm} uses, we train on 44 actions and test on the other 11, using RGB, depth, mmWave, and
WiFi.  For understanding, we use the language-grounded benchmark of HoloLLM under these
cross-action splits.

\noindent\textbf{Unseen actions.}
On either dataset, an action is unseen when no real sensor recording of it is used in training, and no model,
including every baseline, is initialized from weights trained on its recordings.  Pose estimation
and motion prediction thus test \emph{cross-action generalization}, and recognition and QA
test \emph{unseen-action recognition}.

\noindent\textbf{Protocols and metrics.}
All protocols use $T{=}5$ frames at 10\,Hz.  Estimation reports the mean per-joint position error (MPJPE) in absolute
coordinates, Procrustes-aligned MPJPE (PA-MPJPE), and the mean root position error (MRPE)~\cite{moon2019camera}.  Prediction reports the pelvis-relative MPJPE of a single prediction over the next ten
frames (0.1--1\,s, as in mmPred~\cite{fan2026mmpred}).  Captioning reports METEOR~\cite{banerjee2005meteor} and BERTScore~\cite{zhang2020bertscore}.
Action recognition (HAR) and QA report the top-1 accuracy of classification over the unseen
actions~\cite{cao2024mmclip}, where each method picks the candidate with the highest log-likelihood~\cite{alayrac2022flamingo,dai2023instructblip}
after a label-free calibration on the validation
set~\cite{zhao2021calibrate,holtzman2021surface,fei2023mitigating,zhou2024batch,lin2024visualgptscore}.
Paired 95\% confidence intervals
are computed over the test subjects, or over the test sequences for ablations.

\noindent\textbf{Baselines.}
For estimation, we compare with X-Fi~\cite{chen2024xfi} and LEO-Fuse~\cite{aslan2026leofuse}, which fuse any sensor
subset, AWC~\cite{qi2026awc}, which fuses a fixed set of four sensors, and DAPT~\cite{an2025dapt} for
LiDAR.  For understanding, we compare with HoloLLM and with Qwen2.5-VL-7B~\cite{bai2025qwen25vl} on RGB, zero-shot (ZS)
or fine-tuned on the same real samples without simulation (SFT), and all of them receive the same five frames as \method.  For
motion prediction, siMLPe~\cite{guo2023simlpe}, AuxFormer~\cite{xu2023auxformer}, and HumanMAC~\cite{chen2023humanmac}
extrapolate \method's estimated past, and mmPred~\cite{fan2026mmpred} predicts from mmWave.  All baselines except the zero-shot Qwen2.5-VL are trained
on our split, and LEO-Fuse and AWC are reimplemented.

\subsection{Main results}
\label{sec:main}

On the unseen MM-Fi actions (Tab.~\ref{tab:main}), the unified \method model with all five sensors outperforms the models compared on each task.

\noindent\textbf{Estimation.}
With all five sensors, \method has the lowest pose error, 5.5\% below LEO-Fuse and 20\% below X-Fi.  It also outperforms
AWC on the fixed set of four sensors that AWC fuses, and DAPT, a specialist model for LiDAR, on LiDAR alone.

\noindent
\textbf{Understanding.}
\method outperforms HoloLLM and a zero-shot Qwen2.5-VL by more than 25 points in both recognition and QA on the unseen
actions, and both baselines stay below 43\% with or without calibration.
It also writes better captions than HoloLLM, and its BERTScore is higher as well (46.3 \vs 45.3).
Given only RGB, a Qwen2.5-VL fine-tuned on the same timestamped frames and real samples is competitive with \method but
uses 650 visual tokens per observation, whereas we use 20.
  With all five sensors, \method recognizes and
answers more accurately than this Qwen2.5-VL.

\noindent
\textbf{Prediction.}
The prediction of \method is more accurate than those of the compared methods, both on average and 1\,s ahead.  
Given the same mmWave input, it is also more accurate
than mmPred on average.

\subsection{Results on sensor subsets}
\label{sec:subsets}

\begin{table}[t]
\centering
\caption{\textbf{Results by sensor subset}: the same \method model given each subset, against
models given the same subset.  HAR and QA: calibrated top-1 (\%).  Bold: best.  $^\dagger$Our reimplementation.
$^\S$Weighted-sum fusion.}
\label{tab:subsets}
\footnotesize
\setlength{\tabcolsep}{1.4pt}
\begin{tabular}{lccOcOcO}
\toprule
 & \multicolumn{3}{c}{MPJPE$\downarrow$} & \multicolumn{2}{c}{HAR$\uparrow$} & \multicolumn{2}{c}{QA$\uparrow$} \\
\cmidrule(lr){2-4}\cmidrule(lr){5-6}\cmidrule(l){7-8}
Sensors & X-Fi & LEO-Fuse$^\dagger$ & Ours & HoloLLM & Ours & HoloLLM & Ours \\
\midrule
I      & 133.6 & 103.3 & \textbf{90.1} & 31.5 & \textbf{68.2} & 31.1 & \textbf{55.6} \\
D      & 118.3 & 98.5 & \textbf{90.9} & 25.7 & \textbf{59.0} & 23.2 & \textbf{53.4} \\
L      & 163.6 & 93.6 & \textbf{88.9} & 33.2 & \textbf{58.0} & 27.1 & \textbf{57.8} \\
R      & 160.6 & \textbf{153.1} & 154.8 & 25.4 & \textbf{29.8} & 20.1 & \textbf{29.5} \\
W      & 228.8 & 212.6 & \textbf{192.1} & 10.2 & \textbf{20.5} & 11.7 & \textbf{27.1} \\
\midrule
R+W                  & 157.4 & 150.7 & \textbf{144.2} & 13.3$^\S$ & \textbf{27.8} & 14.6$^\S$ & \textbf{32.4} \\
D+L+R+W              & 106.6 & 90.8 & \textbf{87.7} & 21.6$^\S$ & \textbf{60.5} & 20.3$^\S$ & \textbf{64.6} \\
I+D+L+R+W            & 105.1 & 89.0 & \textbf{84.1} & 40.8$^\S$ & \textbf{69.2} & 39.7$^\S$ & \textbf{73.4} \\
\bottomrule
\end{tabular}
\end{table}

Because a missing sensor only turns its joint tokens into recovery seeds, the same \method model handles any of
the 31 nonempty sensor subsets without retraining, so it keeps working when a sensor is switched off. As shown in Tab.~\ref{tab:subsets} and supplementary Sec.~E, it has a lower pose error than X-Fi on all
31 subsets and than LEO-Fuse on 28 of them.
Among single sensors, LiDAR gives the lowest
pose error and RGB the highest recognition accuracy.  With all five sensors, QA accuracy is 15.6 points above the best sensor.

\noindent\textbf{Without RGB.}
\method without RGB still outperforms the multisensory
baselines that use RGB.  From the other four sensors, it estimates pose more accurately than LEO-Fuse and X-Fi with all five sensors,
recognizes unseen actions better than HoloLLM with RGB (60.5\% \vs 42.8\%), and answers more accurately than from RGB
alone (64.6\% \vs 55.6\%).
With a single sensor, its pose error is on average 6.7\% below that of LEO-Fuse, and its recognition and QA accuracy
exceed those of HoloLLM for every sensor.

\begin{figure}[t]
  \centering
  \includegraphics[width=\columnwidth]{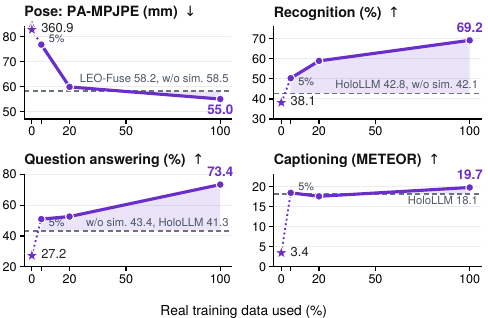}
  \caption{\textbf{Real-data efficiency on unseen actions}, all five sensors.  Stars mark simulation only, off the axis for
  pose, and dashed lines mark models trained on all the real data (HoloLLM with I+R+W).}
  \label{fig:data-efficiency}
\end{figure}

\begin{figure*}[t]
  \centering
  \includegraphics[width=\textwidth]{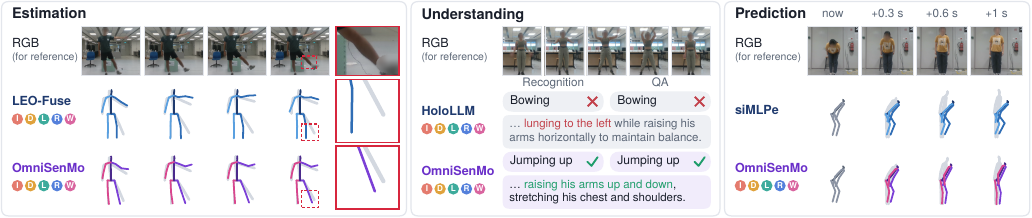}
  \caption{\textbf{Qualitative results on unseen actions.}  Both prediction rows start from \method's estimated past.  Red
  boxes zoom in on the left ankle.  Dots mark each model's sensors, and gray skeletons are the ground truth.  Correct and
  wrong caption phrases are green and red.}
  \label{fig:qual-language}
\end{figure*}

\subsection{Ablations}
\label{sec:ablation}

\begin{table}[t]
\centering
\caption{\textbf{Ablations}: one component of \method is removed or the joint tokens are changed, averaged over five
sensor combinations, one per sensor count (I+D+L+R+W, D+L+R+W, I+D+L, R+W, and I).  In parentheses: change from
\method, gray where its paired 95\% confidence interval includes zero.}
\label{tab:ablation}\label{tab:joint-tokens}
\small
\setlength{\tabcolsep}{3pt}
\begin{tabularx}{\columnwidth}{@{}l*{3}{>{\centering\arraybackslash}X}@{}}
\toprule
Variant & MPJPE$\downarrow$ & HAR$\uparrow$ & QA$\uparrow$ \\
\midrule
\oursrow\method & 97.8 & 59.8 & 60.0 \\
\midrule
w/o joint-token attention & 97.8\,{\scriptsize\textcolor{gray}{(0.0)}} & 38.8\,{\scriptsize($\downarrow$21.0)} & 37.9\,{\scriptsize($\downarrow$22.1)} \\
Pooled tokens & 101.2\,{\scriptsize($\uparrow$3.4)} & 54.6\,{\scriptsize($\downarrow$5.2)} & 53.8\,{\scriptsize($\downarrow$6.2)} \\
w/o joint correspondence & 105.6\,{\scriptsize($\uparrow$7.8)} & 43.6\,{\scriptsize($\downarrow$16.2)} & 47.5\,{\scriptsize($\downarrow$12.5)} \\
w/o recovery & 97.8\,{\scriptsize\textcolor{gray}{(0.0)}} & 58.4\,{\scriptsize\textcolor{gray}{($\downarrow$1.4)}} & 49.1\,{\scriptsize($\downarrow$10.9)} \\
\midrule
w/o simulation & 101.6\,{\scriptsize($\uparrow$3.8)} & 37.1\,{\scriptsize($\downarrow$22.7)} & 38.5\,{\scriptsize($\downarrow$21.5)} \\
w/o VLM teacher & 97.8\,{\scriptsize\textcolor{gray}{(0.0)}} & 56.5\,{\scriptsize($\downarrow$3.3)} & 51.3\,{\scriptsize($\downarrow$8.7)} \\
\midrule
Single protocol & 97.8\,{\scriptsize\textcolor{gray}{(0.0)}} & 49.8\,{\scriptsize($\downarrow$10.0)} & 43.5\,{\scriptsize($\downarrow$16.5)} \\
\bottomrule
\end{tabularx}
\end{table}

Each ablation in Tab.~\ref{tab:ablation} removes one component and is retrained.  Simulation and the joint tokens have
the largest effects, on understanding in particular.  Simulation pairs the sensors with far more actions than the real
data contain, and the joint tokens carry what the encoders learn from it into the language model, so removing either
lowers recognition and QA by more than 20 points.
Except without simulation, with pooled tokens, and without joint correspondence, the variants leave
pose nearly unchanged, since they keep the encoder architecture and its simulated training data.

\noindent\textbf{Skeleton-grounded joint tokens.}
\emph{w/o joint-token attention} retrains the model so that understanding cannot attend to the joint tokens, fused
tokens, or per-frame human queries, while the holistic and visual tokens remain available.
\emph{Pooled tokens} replaces the joint tokens with generic ones.  It still gives $J$ tokens per frame, but they are
formed from the backbone features of the frame after global average pooling, as in CLIP- and ViT-style encoders, and
none of them is tied to a joint.
\emph{w/o joint correspondence} keeps $J$ tokens that attend to local features but ties none of them to a joint and
fuses the sensors without correspondence.  Both variants are retrained from the encoders onward.  Joint tokens outperform these
generic pooled tokens trained on the same simulated data, in both estimation and understanding, and the pose gap grows
from 3.4\,mm on average to 13.2\,mm with mmWave and WiFi alone.
Removing only the correspondence, which leaves the tokens of different sensors spatially misaligned, raises the pose
error by 7.8\,mm on average and lowers recognition and QA by 16.2
and 12.5 points, so the joint tokens help mainly through their correspondence to the joints rather than through their
number.
\emph{w/o recovery} drops the recovery losses, so the seeds of a masked sensor are not trained to recover it.  QA drops
by 10.9 points on average and by 19.5 with all sensors.  Notably, masked sensor recovery improves QA even when no sensor is
missing.

\noindent\textbf{Multisensory sim-to-real pretraining.}
\emph{w/o simulation} keeps Stage~I but skips Stage~II, with the encoders starting from ImageNet or random weights.  We
also repeat real-data adaptation and Stage~III on 5\% and 20\% of the real training sequences, as shown in
Fig.~\ref{fig:data-efficiency}.  The first 20\% gives most of the gain over simulation alone.  With 5\%, recognition
and QA already exceed HoloLLM and \method w/o simulation trained on all the real data.  Simulation thus mitigates the
scarcity of paired real data.
\emph{w/o VLM teacher} does not distill the frozen VLM's RGB features into the non-RGB encoders or the RGB recovery
seeds, while RGB remains an input.  The VLM teacher mainly benefits QA on the unseen actions, which drops by 8.7 points
on average without it and by 8.6 without RGB.

\noindent\textbf{MGO.}
\emph{Single protocol} trains and tests understanding on the dense window, five contiguous frames at the center of the
annotated interval.  Recognition and QA both drop, and with all sensors, recognition of both kicks falls to 1.6\%,
since a kick is recognizable only over its whole course.  Linear probes on the joint tokens, averaged over the five
sensors and evaluated on validation windows of the training actions, decode the pose more accurately from dense windows
(47.5 \vs 52.3\,mm MPJPE) and the action more accurately from sparse samples (100\% \vs 89.5\%, supplementary
Sec.~F.5).  Matching the frames to each task's temporal granularity thus lets the same model serve both estimation and
understanding.

\subsection{Cross-dataset validation on XRF55}
\label{sec:xrf55}

\begin{table}[t]
\centering
\caption{\textbf{Results on unseen XRF55 actions}.  HAR and QA: calibrated top-1 (\%), uncalibrated in parentheses.  Bold: best in each block.
$^\S$Weighted-sum fusion.}
\label{tab:xrf55}
\footnotesize
\setlength{\tabcolsep}{3pt}
\begin{tabular}{llccc}
\toprule
Sensors & Method & HAR$\uparrow$ & QA$\uparrow$ & METEOR$\uparrow$ \\
\midrule
\multirow{4}{*}{I} & Qwen2.5-VL-ZS & 65.7\raw{61.0} & \textbf{66.3}\raw{59.7} & 7.3 \\
 & Qwen2.5-VL-SFT & 64.8\raw{61.9} & 65.9\raw{63.2} & \textbf{19.6} \\
 & HoloLLM~\cite{zhou2025holollm} & 24.6\raw{17.5} & 25.2\raw{14.2} & 18.2 \\
\oursrow\nobg & \method & \textbf{65.8}\raw{59.8} & 64.2\raw{59.3} & 19.5 \\
\midrule
 & HoloLLM & \textbf{26.7}\raw{18.0} & 25.2\raw{16.5} & \textbf{18.5} \\
\oursrow\nobg\multirow{-2}{*}{D} & \method & 26.4\raw{22.1} & \textbf{28.4}\raw{26.3} & \textbf{18.5} \\
\midrule
 & HoloLLM & 13.1\raw{16.4} & 12.9\raw{14.8} & \textbf{17.4} \\
\oursrow\nobg\multirow{-2}{*}{W} & \method & \textbf{21.6}\raw{15.6} & \textbf{19.9}\raw{17.3} & \textbf{17.4} \\
\midrule
 & HoloLLM$^\S$ & 20.3\raw{19.7} & 22.5\raw{16.0} & 18.3 \\
\oursrow\nobg\multirow{-2}{*}{I+D+W} & \method & \textbf{63.4}\raw{56.5} & \textbf{63.2}\raw{56.1} & \textbf{18.5} \\
\midrule
\oursrow\nobg I+D+R+W & \method & 64.3\raw{56.6} & 63.9\raw{57.0} & 18.4 \\
\bottomrule
\end{tabular}
\end{table}

To validate the pipeline on new hardware and unseen actions, we re-simulate mmWave and WiFi with the XRF55 hardware parameters,
since its radar gives heatmaps instead of point clouds, and adapt the model to the real clips of the training actions.
Given the same sensors (Tab.~\ref{tab:xrf55}), the model recognizes the unseen actions far better than HoloLLM for every
sensor subset except depth alone, where the two are on par.
On the same RGB input, the multisensory \method is competitive with Qwen2.5-VL.

\subsection{Qualitative results}
\label{sec:qualitative}

Figure~\ref{fig:qual-language} illustrates the results of Sec.~\ref{sec:main} on unseen actions with all five sensors.
In estimation, \method follows the extended leg that LEO-Fuse misses, with about half the error of LEO-Fuse at the left
ankle.  In understanding, it names a jump that HoloLLM mistakes for bowing, and its caption describes the arm movement,
while that of HoloLLM describes a lunge.  In prediction, siMLPe keeps the subject bent forward from the same estimated
past, whereas \method predicts the rise from the bow.

\section{Conclusion}
\label{sec:conclusion}

We presented \method, a unified multisensory language model that estimates, understands, and predicts human motion from
any subset of RGB, depth, LiDAR, mmWave, and WiFi. With the human skeleton as a common reference, its joint tokens address the
semantic inconsistency and spatial misalignment of heterogeneous sensors and fuse any sensor subset joint by joint in the motion vocabulary that the language model learned from motion capture and text.
Multi-Granularity Observation matches the frames to each task's temporal granularity, and pretraining on simulated
sensors covers actions that real sensor data lack. On the cross-action split of MM-Fi, the same model with all five
sensors outperforms the models compared on each task, and in a cross-dataset validation on XRF55, it recognizes unseen
actions far better than HoloLLM. These results make the skeleton a bridge between human sensing and language models.

\section*{LLM Usage Statement}
We used Claude (Anthropic) as a coding and writing assistant to write and debug the training and evaluation code,
launch experiments and compute metrics from their stored outputs, make figures and tables, search and summarize related
papers, discuss design options for the method and experiments, and draft and edit the text, including grammar.  We also
used OpenAI Codex for code.  The house, robot, and person illustrations in Fig.~\ref{fig:teaser} were generated with
OpenAI's image generation.  No generative AI tool was used to produce experimental data or results, and Qwen2.5-VL and
the vision--language teacher in \method are components of the method, not writing aids.  The authors made or approved
all research decisions, verified every reported number against the stored outputs, and take full responsibility for
the content.

\section*{Acknowledgments}
This work used Jetstream2 at Indiana University and Delta at the National Center for Supercomputing Applications
(NCSA) through allocation CIS262336 from the Advanced Cyberinfrastructure Coordination Ecosystem: Services \&
Support (ACCESS) program, which is supported by U.S. National Science Foundation grants \#2138259, \#2138286,
\#2138307, \#2137603, and \#2138296. The work of Pu (Perry) Wang is exclusively supported by MERL.

{
    \small
    \bibliographystyle{ieeenat_fullname}
    \bibliography{references}
}
\clearpage
\includepdf[pages=-,pagecommand={\thispagestyle{plain}}]{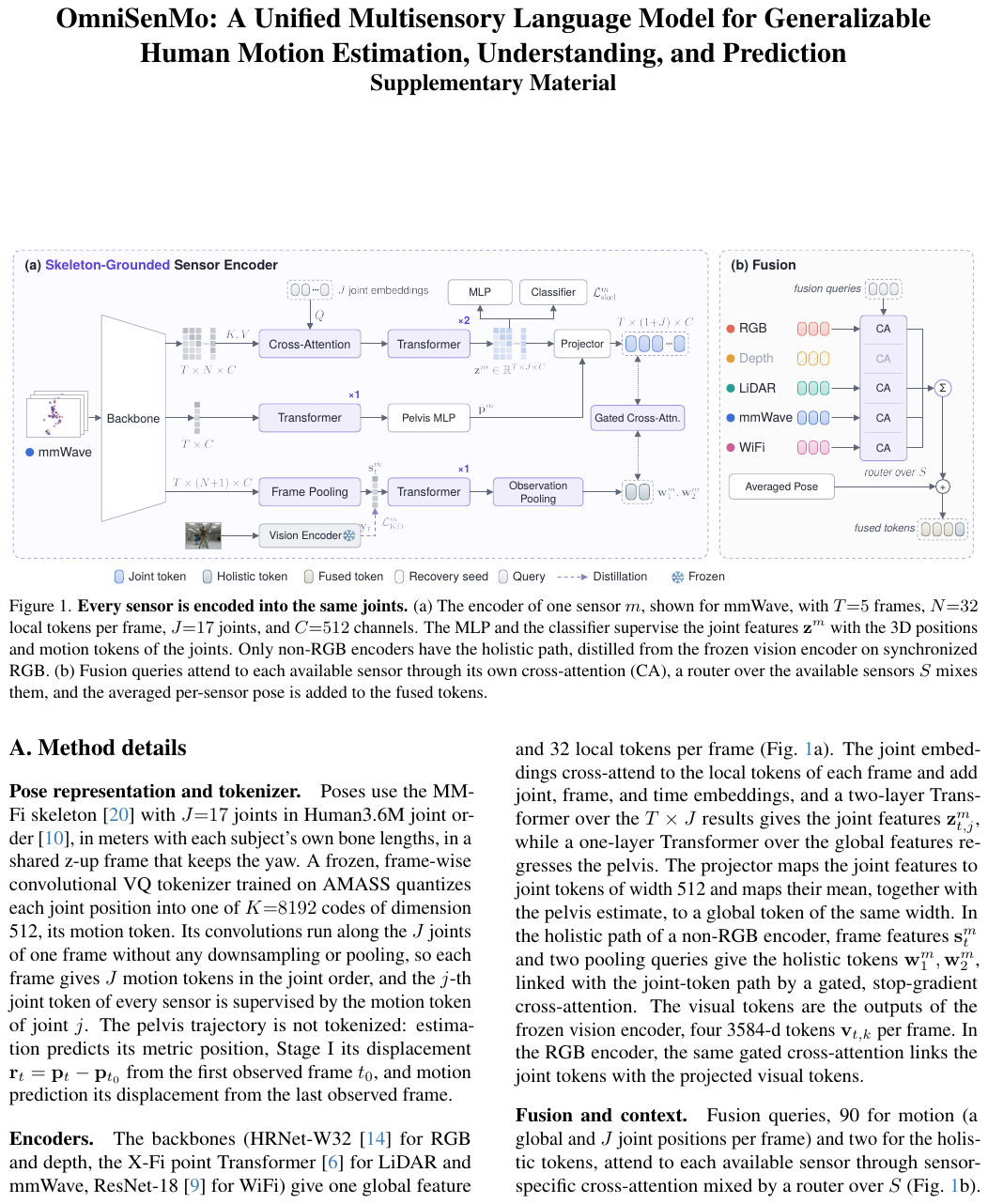}
\end{document}